\documentclass[a4paper,11pt]{amsart}
\pdfoutput=1

\usepackage{hyperref}
\usepackage{amsmath,amsthm}
\usepackage[foot]{amsaddr}
\usepackage{booktabs} 
\usepackage{multirow} 
\usepackage{placeins} 
\usepackage{xcolor}
\usepackage{lscape} 

\calclayout

\title{Territory Design for Dynamic Multi-Period Vehicle Routing Problem with Time Windows}

\author{Hern\'an Lespay$^1$}
\author{Karol Suchan$^{2,3}$}
\address{$^1$Universidad Adolfo Ib\'a\~nez, Av. Diagonal las Torres 2640, Santiago 7941169, Chile}
\address{$^2$Universidad Diego Portales, Av. Ej\'ercito Libertador 441, Santiago 8370191, Chile}
\address{$^3$AGH University of Science and Technology, al. A. Mickiewicza 30, Krakow 30-059, Poland}
\email{hlespay@alumnos.uai.cl, karol.suchan@mail.udp.cl}

\begin{document}

\begin{abstract}\small
This study introduces the Territory Design for Dynamic Multi-Period Vehicle Routing Problem with Time Windows (TD-DMPVRPTW), motivated by a real-world application at a food company's distribution center. This problem deals with the design of contiguous and compact territories for delivery of orders from a depot to a set of customers, with time windows, over a multi-period planning horizon. Customers and their demands vary dynamically over time. The problem is modeled as a mixed-integer linear program (MILP) and solved by a proposed heuristic. The heuristic solutions are compared with the proposed MILP solutions on a set of small artificial instances and the food company's solutions on a set of real-world instances. Computational results show that the proposed algorithm can yield high-quality solutions within moderate running times.
\end{abstract}

\keywords{territory design, vehicle routing, multi-period routing, distribution logistics, heuristics}

\subjclass[2010]{90B06, 90B10, 90B50, 90B90, 68T20, 68W05}

\maketitle

\section{Introduction}
Territory Design (TD) problems consist in grouping small geographic areas, called basic units, into larger zones, called territories, which must meet a set of planning criteria depending on the specific application context. The planning criteria most often used correspond to balance, contiguity, and compactness of territories. The balance criterion consists in designing territories of equitable size. The contiguity or connectivity criterion ensures that it is possible to move between any pair of basic units within one territory without crossing any other. And the criterion of compactness corresponds to designing geographically compact territories, i.e., with a dense structure of basic units closely packed together. The contiguity and compactness criteria are widely used in Territory Design because they generally reduce distances and travel times within a territory, directly benefiting the company's operational costs. The balance criterion is related to fairness, for example, with respect to the drivers that execute the deliveries.

The TD problem studied in this paper is motivated by a real-world application at a food distribution company in Chile. This company is one of the leading national producers and marketers of meat and poultry. The company has seventeen distribution centers throughout the national territory. In this paper, we focus on just one of them. However, this study could be replicated at any of the company's distribution centers since their operations are quite similar throughout the country. The company has large sales volumes nationwide, and the costs associated with distribution activities represent an essential component of their total expenditures. Therefore, a reduction of 5 or 10 percent in the routing costs would mean a significant improvement for the company.

In this work, the company's customers correspond to supermarkets, restaurants, wholesalers, and small convenience stores spread throughout a region of Chile. The set of customers changes, on average, by 10\% from one month to another. This customer variability is mainly due to the low frequency and irregularity of orders from small convenience stores, which represent 60\% of the total number of customers. Also, the set of customers is characterized by high differences in service times. Even under perfect conditions, the time needed to unload the delivery can differ by about an hour from one customer to another. Additionally, each customer requires that their orders be delivered within a certain time-window. The company has a fleet of vehicles dispatched from the distribution center to their respective customers. Each vehicle has a fixed subset of customers assigned to it. On each day, the vehicles serve their corresponding customers.

Due to the high variability in the activity of customers (if they make an order for the particular day or not), the quantities they order, and the service times, the company currently experiences significant problems in the fulfillment process during periods of high activity: the customers' time windows may not be respected and, in some cases, the delivery even has to be postponed to another day. Moreover, the company experiences truck hijacking cases, where criminals take control of the truck to steal the goods and abandon the vehicle in some remote place. To improve the drivers' security and diminish the losses, it would be valuable to implement an early warning system to raise an alert each time an atypical behavior of a truck is detected (see, for example, \cite{oliveira2015}). Consequently, the company needs to develop a decision support system that would improve the planning, scheduling, and control of products' distribution to its customers, seeking the right balance between operational costs, service quality, and security.

We propose an algorithm to output customer-to-driver assignments that correspond to contiguous and compact geographical zones in this work. Such a characteristic is very valuable for the company's management, for the ease of establishing service contracts with third-party transportation companies based on territories and not on particular sets of clients. Moreover, it enables the definition of geofences - to trigger an alarm each time the truck leaves its corresponding zone (based on a real-time stream of GPS data from all the trucks). Our algorithm represents a useful tool to determine the best way to serve and respond quickly to new customers, standardize service quality, improve security, and efficiently allocate customers among their different drivers.

The remainder of this paper is organized as follows. Section 2 introduces the literature review of TD for products' distribution.  Section 3 presents a description of the problem. Section 4 describes the solution framework. Section 5 reports the results of computational experiments. Finally, section 6 presents the conclusions.

\section{Literature Review}

TD has been addressed in various application contexts. There is extensive literature on political, school, healthcare services, police, sales, and marketing districts. For a general review on TD with different applications we refer the reader to \cite{kalcsics2005,zoltners2005,ricca2013,rios2020}. The present review focuses on works that consider territory design with some vehicle routing aspects, particularly on the design of territories for product distribution.

In \cite{jarrah2012}, it is argued that making routing decisions based on territory design leads companies to more efficient distribution operations, generating several benefits for the drivers and the company. With a stable work area, the drivers are comfortable with their routes, leading to better reliability and shorter service times. On the other hand, it allows the customers to become familiar with the drivers, increasing customer satisfaction. Similarly, in \cite{schneider2015}, it is pointed out that maintaining a fixed driver that regularly visits a constant set of clients allows him to become familiar with his territory and his customers. This situation contributes to the driver serving customers more efficiently because experienced drivers know shortcuts, anticipate congestion problems, and find parking spots more quickly, reducing travel and customer service times.

Territory design decisions are made at the strategic level, and routing decisions are made at the operational level. At the strategic level, works on territory design with vehicle routing considerations share some common requirements. Each customer must be served by only one vehicle; thus, each basic unit must be assigned to a single territory. The composition of territories has to allow for daily routing that respects the vehicles' capacities and operational time limits. Contiguity and compactness are common characteristics as well, not only because they heuristically contribute to reducing the total travel times, but they also simplify the management of relations with truckers and clients (that may come and go, causing a need for adaptations in the distribution process). Otherwise, solving VRP for each day independently would lead to routing plans of short total distance and travel time, however, at the cost of frequently changing vehicle routes and customer-to-driver assignments. Such variability can negatively impact drivers' productivity and job satisfaction, diminish client satisfaction, and cause a great managerial burden, for example, in cases of distribution executed by third party transportation companies.

At the operational level, routing decisions change depending on the kind of VRP which is approached. Some works consider VRP for deterministic location customers in a single-period \cite{lin2017,jarrah2012,jarrah2011,haugland2007} (in \cite{haugland2007}, additionally a variant in which the demand is stochastic, but and not null, for all customers is considered). Other works consider VRP for dynamic-deterministic location customers in multiple-periods \cite{lei2015}. Unlike the former, dynamic-deterministic location customers mean that the active customers set varies dynamically over the planning horizon. However, complete information on their behavior is available in advance.

In \cite{lin2017}, the Meals-On-Wheels Service Districting (MOWSD) problem was addressed, motivated by an application for Home Health Care (HHC) services. This problem aims to find the minimum number of districts to cover all customers, satisfying capacity, and work time limitations in each district. The districts' compactness was modeled as a constraint limiting the maximum walking/driving duration between any two basic units in the same district. A MILP formulation was presented, and a greedy heuristic was developed to solve the problem. Experiments were performed using a real instance with road distances from Google Maps. The results showed that the greedy heuristic could find solutions as good as those found by the Gurobi Optimizer working on the MILP formulation. Moreover, the proposed heuristic found territories that were more compact than the ones used by HHC, defined by manual planning. 

In \cite{jarrah2011,jarrah2012}, a problem of partitioning a local service region into non-overlapping and contiguous work areas, in which pickups and deliveries are made, was addressed. The objective is to find the least number of work areas, where each satisfies some shape constraints and admits a route that satisfies vehicle capacity and total travel time restrictions. The working areas' compactness is reached by relying on the bounding rectangles and restricting their aspect ratio to lie within a predefined range. The problem was modeled using a set-covering formulation and was solved with an adaptive column generation heuristic. Additionally, in \cite{jarrah2012}, a set of geometric constraints that allows for the implicit generation of clusters was introduced. Moreover, a metaheuristic was developed for obtaining high-quality solutions to large-scale set covering problems. The algorithms were tested on instances provided by an international express package carrier considering Euclidean distances. The results showed that vehicle reductions averaging 7.6\% could be reached by adopting the proposed solutions.

In \cite{haugland2007}, designing districts for vehicle routing problems with static customers and stochastic demands was addressed. The demand is not null for all customers throughout the planning horizon, but its precise value is observed only after the districts have been defined. Then, a travel plan must be produced within each district, solving a deterministic TSP. As a result, some vehicle routes may violate the capacity constraints. Thus, breaks must be positioned such that the total demand realized in each subtour does not exceed the vehicle capacity (thus transforming TSP into VRP). The problem was modeled as a two-stage stochastic program. In the first stage, demand is considered a stochastic variable, and the districting decisions are made based on the expected route length. In contrast, in the second stage, one VRP is solved for each district deterministically. The compactness of the districts is reached by selecting the closest customers. A Tabu Search metaheuristic and a multi-start heuristic were developed to solve the problem. Experiments were performed on modified literature instances for VRP and considering Euclidean distances. The results show that the Tabu Search metaheuristic has a better performance than the multi-start heuristic.

In \cite{lei2015}, the Multiple Traveling Salesmen and Districting Problem with Multi-Periods and Multi-Depots was introduced. The problem involves designing contiguous districts and subdistricts for multiple traveling salesman problems (MTSP) with dynamic customers over several periods. The objective is to optimize the number of districts, their compactness and dissimilarity, and the equity of the salesmen's profits. Here, the total travel time limit should be respected; otherwise, an overtime cost is incurred. An adaptive large neighborhood search metaheuristic (ALNS) was developed for the problem. Experiments were conducted on modified Solomon and Gehring \& Homberger instances (demand and time windows were not used) to assess the proposed algorithm's quality, considering Euclidean distances. The experiments presented focused on comparing the solutions obtained by a construction heuristic proposed to generate a feasible initial solution with the ones obtained through ALNS. Experiments confirmed the proposed metaheuristic's effectiveness.

Unlike the previously cited works, there exist studies that consider solving VRP for stochastic location customers for multiple-periods \cite{lei2012,lei2016}. Stochastic location customers mean that the active customers set is associated with uncertainty in customers' number and location over the planning horizon. The precise information is available only after the districts have been created. In these studies, a TSP is solved for each district, with a maximum route length restriction. If the maximum route length restriction is violated, then a penalty cost is incurred.

In \cite{lei2012}, the Vehicle Routing and Districting Problem with Stochastic Customers (VRDPSC) in multiple-periods was approached. It consists in designing contiguous districts for a vehicle routing problem with stochastic location customers to minimize the number of vehicles used, expected routes cost, and a district compactness measure. Each vehicle route must satisfy capacity and travel time limitations. If the maximal duration of a route is exceeded, an overtime cost is incurred. The problem was modeled as a two-stage stochastic problem. In the first stage, the decisions on the design of districts are made. In the second stage, the expected routing cost is obtained for each district. A large neighborhood search (LNS) heuristic was developed to solve the problem. Like in \cite{lei2015}, experiments were conducted on modified Solomon and Gehring \& Homberger instances to assess the quality of the proposed algorithm, considering Euclidean distances. Similarly, the experiments focused on comparing the solutions obtained by a construction heuristic proposed to generate a feasible initial solution with the LNS heuristic ones. The results confirmed the effectiveness of the LNS heuristic. Later, in \cite{lei2016}, the Multi-objective Dynamic Stochastic Districting and Routing Problem (MDSDRP) was approached. This problem is related to the previous ones (\cite{lei2012,lei2015}). Still, with some differences: (i) the former works focused on a single aggregated objective problem, whereas the latter paper studied a multi-objective problem that minimizes the average number of districts, the average compactness of districts, the average dissimilarity of districts, and the average equity of salesmen's profits, (ii) the customers in the two previous works were respectively stochastic and dynamic, whereas in this problem, one part of the customers is dynamic (their activity varies throughout the time horizon, but the demand is known in advance) and the other is of stochastic location, and (iii) in the previous two studies, the authors used large neighborhood search heuristics, whereas a multi-objective evolutionary algorithm (MOEA) was developed in this study. The algorithm was tested on randomly generated instances using Euclidean distances. Regular and stochastic customers were uniformly distributed in a square. The proposed algorithm was compared with two state-of-the-art MOEAs because no previous computational results were available. Computational results confirmed the superiority and effectiveness of the proposed algorithm. 

In the cited works, when the demand is not known at the time of district definition, like in \cite{haugland2007} or, when the number and location of customers change in every period, like in \cite{lei2012,lei2015,lei2016}, the territories are created with incomplete information about customers or their demand. Then, the feasibility of routing plans must be evaluated. However, computing the expected tour length of a given district requires solving several routing problems, which is expensive in terms of computational resources. To avoid the computational burden, heuristics for territory design were guided by approximate solutions, determining upper bounds for each district's routing subproblems \cite{haugland2007,lei2012,lei2015,lei2016}. Particularly, in \cite{haugland2007}, the expected tour length is approximated by heuristically determining upper bounds for the routing plan and, in \cite{lei2012,lei2015,lei2016}, the Beardwood-Halton-Hammersley formula is used.

\section{The Problem}

\begin{table}[th!]
\caption{Summary of notation}\label{tab:notation}
\begin{tabular}{p{1cm} p{15.5cm}}
\toprule
\multicolumn{2}{l}{\textbf{Sets}}\\
$V$ & Set of nodes (customers, basic units).\\
$A$ & Set of arcs (adjacency of basic units).\\
$K$ & Set of territories (vehicles).\\
$D$ & Set of days.\\
$N(i)$ & Set of neighbors of $i\in V$.\\
$V_k$ & Set of customers of territory $k\in K$, $V_k \subset V$.\\
\multicolumn{2}{l}{\textbf{Parameters}}\\
$n$ & Number of nodes (customers).\\
$q_{i,d}$ &Demand of customer $i\in V{\setminus} \{0\}$ on day $d\in D$.\\
$g_i$ & Service time of customer $i\in V{\setminus} \{0\}$.\\
$o_{i,d}$ & Equal to $1$ if customer $i\in V{\setminus} \{0\}$ requires service on day $d$ ($q_{i,d} > 0$), and $0$ otherwise.\\
$[a_i,b_i]$ & Time window of customer $i\in V$, where $a_i$ and $b_i$ are its lower and upper bound, respectively.\\
$t_{i,j}$ & Travel time between customers $i$ and $j$, for $(i,j)\in A$.\\
$c$ & Vehicle capacity (the same for all $k\in K$).\\
$h$ & Working day hours.\\
$\hat{f}_i$ & Perimeter of basic unit $i \in V$.\\
$f_{i,j}$ & Boundary length between basic units $i \in V$ and $j \in V$, $i \neq j$.\\
$m_i$ & Square root of area of basic unit $i \in V$.\\ 
$F$ & Compactness bounding parameter. \\
\multicolumn{2}{l}{\textbf{Binary variables}}\\
$x_{i,j,k,d}$ &Equal to $1$ if arc $(i,j)\in A$ is traversed by vehicle $k\in K$ on day $d\in D$, and $0$ otherwise.\\
$y_{i,k,d}$ &Equal to $1$ if customer $i\in V{\setminus}\{0\}$ is assigned to territory $k\in K$ on day $d\in D$, and $0$ otherwise.\\
$z_{k}$ &Equal to $1$ if vehicle $k\in K$ is used on any day of the planning horizon, and $0$ otherwise.\\
$\hat{z}_{i,k}$ &Equal to $1$ if customer $i\in V{\setminus}\{0\}$ is assigned to territory $k\in K$, and $0$ otherwise.\\
$\bar{z}_{i,j,k}$ & Equal to $1$ if customers $i \in V{\setminus}\{0\}$ and $j \in V{\setminus}\{0\}$ are in territory $k\in K$, and $0$ otherwise.\\
$w_{i,k}$ &Equal to $1$ if basic unit $i\in V{\setminus}\{0\}$ is choosen the sink of territory $k\in K$, and $0$ otherwise.\\
\multicolumn{2}{l}{\textbf{Continuous variables} }\\
$e_{j,d}$ & Waiting time if the vehicle arrives at customer $j\in V{\setminus} \{0\}$ on day $d\in D$ earlier than $a_j$, and $0$ otherwise.\\
$s_{i,d}$ &Service start-time at customer $i\in V{\setminus} \{0\}$ on day $d\in D$. Equal to $0$ if no service is required.\\
$u_{i,j,k}$ &Amount of flow from basic unit $i\in V{\setminus} \{0\}$ to basic unit $j\in V(i)$ in territory $k\in K$.\\
\bottomrule
\end{tabular}
\end{table}

The TD-DMPVRPTW problem is defined on a graph $G=(V,A)$, where $V=\{0,1,...,n\}$. $0$ represents the depot's basic unit, and positive integers represent the basic units of customers. We speak of customers and their corresponding basic units interchangeably. In our instances, each client's basic unit is a convex polygon - the respective cell in the Voronoi diagram of all customers. The graph $G$ describes de topology: two basic units are adjacent if they share a side. Each basic unit has its perimeter $\hat{f}_i$ and the square root of its area $m_i$. Also, for any pair of adjacent basic units, the length of the shared side is denoted by $f_{ij}$. Each arc $(i,j) \in A$ is associated with a travel time $t_{ij}$. In our real-world instances, the travel times are computed using the road network between the respective clients. We assume that the travel times satisfy the triangle inequality.

A territory is a subset of nodes $V_k \subset V$, $k \in K$. Customers in each territory are visited on routes traversed by one of a fleet of homogeneous vehicles in the set $K$. The number of vehicles is not restrictive (i.e., $|K| = |V {\setminus} \{0\}|$). It is required that each customer is assigned to only one territory. Thus, the territories define a partition of $V{\setminus} \{0\}$. Each territory must be contiguous, i.e., for each territory $V_k$, $k \in K$, a vehicle must be capable of moving between any pair of basic units $i\in V_k$ and $j\in V_k$ within territory $V_k$ without crossing any other. The compactness of each territory is assured by upper-limiting its perimeter by $F$ times the sum of the square roots of the areas of the territory's basic units. We define $N(i)$ as the set of customers adjacent to $i$.

The measure of compactness of a territory that we adopt for the heuristic algorithm, the {\em compactness ratio (CR)}, is based on the {\em isoperimetric quotient} $\frac{4 \Pi A}{L^2}$, where $A$ is the area and $L$ is the perimeter (see, for example, \cite{montero2009,deford2019}). In general, there is $\frac{4 \Pi A}{L^2} \leq 1$, with equality if and only if the territory is a circle. Suppose we want to fix a minimum value $\alpha$ for the quotient. So we get $\alpha \leq \frac{4 \Pi A}{L^2}$, which leads to $L^2 \leq \frac{4\Pi}{\alpha}A$. For practical reasons, we take the square roots and get $CR \leq F $, where $CR= \frac{L}{\sqrt{A}} $ and $F=\sqrt{\frac{4\Pi}{\alpha}}$. The inequality holds with $F=2\sqrt{\Pi} \approx 3.5$ only in the case of a circle, otherwise $F$ has to be larger. Notice there is $CR=4$ in case of a square.

In our problem, the area of a territory is computed as the sum of the areas of the basic units that it is composed of. Since dealing with the square root of the sum of the basic units' areas is not straightforward in a MILP model, and our model is used only for benchmarking, we opted for using (the estimation of) the sum of the square roots instead. For the benchmarking we do on small instances, we use the sum of the square roots for both the MILP model and our algorithm, but we keep the ``proper'' square root of the sum when running the algorithm on real-life instances.

We have evaluated several compactness measures, and it was the isoperimetric quotient that best suited our needs. Indeed, one of the characteristics of this measure that is commented on in \cite{deford2019} is that ``this quotient, however, is unstable in the sense that a small perturbation of the shape's boundary can greatly increase its perimeter without significantly affecting its area''. However, for the company under study, in particular, in order to establish the geofences, the priority to ``almost convex'' shapes implied by this measure is an advantage that complements the ``compactness'' (oblivious to the form of the boundary).

Each vehicle $k \in K$, with given capacity $c$, is located at the depot, from where it departs at time $0$ and where it must return before time $h$. The planing horizon involves $|D|$ days, where $D$ is the set of days. On each day $d \in D$, each customer $i \in V {\setminus} \{0\}$ has a demand $q_{i,d}$ and service time $g_i$. We assume that $q_{i,d} \leq c$ and it is not possible to perform split deliveries (each client is assigned to only one territory). We use auxiliary parameters $o_{i,d}$ equal to $1$ if customer $i$ requires service on day $d$ ($q_{i,d} > 0$), and equal $0$ otherwise.

The vehicles have to respect the time windows: the service start-time of customer $i \in V {\setminus} \{0\}$ can vary freely within the interval $[a_i,b_i]$ defined as the customer's time window, where $a_i$ and $b_i$ are its lower and upper bound, respectively. We only require the visit start-time to fall within the time window, not the sum of start- and service-time. A vehicle is allowed to complete the delivery after the time window closes. For ease of presentation, we identify each vehicle $k$ with the set of customers assigned to it and the territory comprised by the union of the respective basic units (we will speak of a vehicle, a territory, and the respective set of customers interchangeably).

A solution $R$ of a TD-DMPVRPTW instance is determined by $|K|$ territories: each one of them, $k\in K$, with an independent route (the order of visits for frequent customers is not preserved from one day to another), $r_{k,d}$, for every day, $d\in D$, when some of the customers in $k$ require service. The objective is to minimize the number of non-empty territories, such that i) the total quantity carried by each vehicle on each day does not exceed the capacity $c$, ii) the time windows of all customers and the depot are respected, iii) exactly one driver serves each customer over the planning horizon, and iv) the contiguity and compactness restrictions are satisfied. The customers are not visited on the days when their demand is $0$. Empty territories are not assigned to drivers, so the number of vehicles used over the planning horizon is decided in the model.

We present a MILP formulation for the TD-DMPVRPTW. The model uses the binary decision variables $x_{i,j,k,d}$ that equal $1$ if arc $(i,j)$ is traversed by vehicle $k$ on day $d$, and $0$ otherwise. The binary decision variables $y_{i,k,d}$ equal $1$ if customer $i$ is assigned to territory $k$ on day $d$, and $0$ otherwise. The binary decision variables $z_{k}$ equal $1$ if vehicle $k$ is used in any day of the planning horizon, and $0$ otherwise. The binary decision variables $\hat{z}_{i,k}$ equal $1$ if customer $i$ is assigned to territory $k$, and $0$ otherwise. The binary decision variables $\bar{z}_{i,j,k}$ equal $1$ if customers $i$ and $j$ are assigned to the same territory $k$, and $0$ otherwise. The binary decision variables $w_{i,k}$ equal $1$ if basic unit $i$ is choosen as the sink of the territory $k$, and $0$ otherwise. The non-negative continuous decision variables $s_{i,d}$ equal the service start-time at customer $i$ on day $d$, and $0$ if no service for customer $i$ is required on day $d$. We define the service start-time at customer $j$ on day $d$ as $s_{j,d} = s_{i,d} + g_i + t_{i,j} + e_{j,d}$, where $e_{j,d}$ is equal to the waiting time if the vehicle arrives at customer $j$ on day $d$ earlier than $a_j$, and $0$ otherwise. Finally, the non-negative continuous decision variables $u_{i,j,k}$ determine the amount of flow from basic unit $i$ to basic unit $j$ in territory $k$ (used to assure contiguity of territories). We summarize all the notation of the problem in Table \ref{tab:notation}.
\newpage
\begin{align}
&Minimize \quad  \sum_{k \in K} z_k 
\end{align}
subject to:\\
\begin{align}
&\sum_{k \in K} y_{i,k,d} = o_{i,d}    & \forall i\in V {\setminus}\{0\}, d \in D \\
&\sum_{i\in V {\setminus}\{0\}} q_{i,d} y_{i,k,d} \leq c    &\forall k\in K, d \in D\\
& z_k \geq \hat{z}_{i,k}  	& \forall i\in V {\setminus}\{0\}, k \in K\\
&\hat{z}_{i,k} \geq y_{i,k,d} & \forall i\in V {\setminus}\{0\}, k \in K, d \in D\\
&\hat{z}_{i,k} \leq \sum_{d\in D}y_{i,k,d} & \forall i\in V {\setminus}\{0\}, k \in K\\
&\bar{z}_{i,j,k} \leq \hat{z}_{i,k} &\forall i,j \in V {\setminus}\{0\}, k \in K\\
&\bar{z}_{i,j,k} \leq \hat{z}_{j,k} &\forall i,j \in V {\setminus}\{0\}, k \in K\\
&\hat{z}_{i,k} + \hat{z}_{j,k} \leq 1 + \bar{z}_{i,j,k} &\forall i, j \in V {\setminus}\{0\}, k \in K\\
&\sum_{i \in V {\setminus}\{0\}} \hat{f}_{i} \hat{z}_{i,k} - \sum_{i \in V {\setminus}\{0\}} \sum_{j \in V {\setminus}\{0\}: j \neq i} f_{i,j} \bar{z}_{i,j,k} \leq F \sum_{i \in V {\setminus}\{0\}} m_i \hat{z}_{i,k} &k \in K\\
&\sum_{j\in N(i)} u_{i,j,k} - \sum_{j\in N(i)} u_{j,i,k} \geq \hat{z}_{i,k} - (|V|-1)w_{i,k} & \forall i\in V {\setminus}\{0\}, k \in K\\
&\sum_{i\in V{\setminus} \{0\}} w_{i,k} = 1 & \forall k \in K\\
&\sum_{j\in N(i)} u_{j,i,k} \leq (|V|-1)\hat{z}_{i,k} & \forall i\in V {\setminus}\{0\}, k \in K\\
&\sum_{i \in V{\setminus} \{j\}} x_{i,j,k,d} = \sum_{i \in V{\setminus} \{j\}} x_{j,i,k,d} = y_{j,k,d}  &\forall j\in V {\setminus} \{0\}, k\in K, d \in D\\
&\sum_{j \in V {\setminus} \{0\}} x_{0,j,k,d}= \sum_{i \in V {\setminus} \{0\}} x_{i,0,k,d} = z_k  & \forall k \in K, d\in D
\end{align}
\begin{align}
&o_{i,\alpha} + o_{i,\beta} - 2 \leq y_{i,k,\alpha} - y_{i,k,\beta}  & \forall i\in V {\setminus}\{0\}, k\in K, \alpha,\beta \in D, \alpha \neq \beta\\
&s_{i,d} + x_{i,j,k,d}(g_i + t_{i,j} + e_{j,d}) - (1-x_{i,j,k,d})T \leq s_{j,d}  &\forall i\in V, j\in V {\setminus}\{0\}, i \neq j, k\in K, d \in D\\
&s_{i,d} + x_{i,j,k,d}(g_i + t_{i,j} + e_{j,d}) + (1-x_{i,j,k,d})T \geq s_{j,d}  &\forall i\in V, j\in V {\setminus}\{0\},i \neq j, k\in K, d \in D\\
x&s_{i,d} + g_i + t_{i,0} \leq h  & \forall i\in V {\setminus}\{0\}, d \in D\\
&a_i o_{i,d}\leq s_{i,d} \leq b_i o_{i,d} & \forall i \in V, d \in D\\
&z_k \in \{0,1\}  &\forall k \in K\\
&\hat{z}_{i,k} \in \{0,1\}  &\forall i \in V, \forall k \in K\\
&\bar{z}_{i,j,k} \in \{0,1\}  &\forall i,j \in V, \forall k \in K\\
&w_{i,k} \in \{0,1\}  &\forall i \in V{\setminus}\{0\}, \forall k \in K\\
&x_{i,j,k,d} \in \{0,1\}   & \forall i,j \in V,i \neq j, k\in K, d \in D\\
&y_{i,k,d} \in \{0,1\}   & \forall i \in V, k\in K, d \in D\\
&s_{i,d} \geq 0   & \forall i \in V, d \in D\\
&u_{i,j,k} \geq 0  & \forall i\in V{\setminus}\{0\}, j\in V(i), k\in K
\end{align}

The objective function $(1)$ minimizes the number of territories. Constraints $(2)$ guarantee that each customer is attended on each day they require service, and inequalities $(3)$ make sure that the vehicle capacity is not exceeded. Constraints $(4)$ force that if customer $i$ is served by vehicle $k$, then vehicle $k$ must be used during the planning horizon.

Constraints $(5)$ and $(6)$ enforce that if customer $i$ is served by vehicle $k$ on day $d$, then customer $i$ has to be assigned to vehicle $k$. Otherwise, the customer and the vehicle are not related. Constraints $(7) - (9)$ ensure that if customers $i$ and $j$ are in the same territory $k$, then $\bar{z}_{i,j,k}$ takes value 1, otherwise, the customers are in different territories and $\bar{z}_{i,j,k}$ takes value 0. Constraints $(10)$ ensure that the respective territory's perimeter is at most $F$ times the sum of the square roots of the areas of the basic units that it is composed of, where $F$ is a parameter of the problem.

Constraints $(11)$-$(13)$ are the contiguity constraints. These constraints were proposed in \cite{shirabe2005} and guarantee the connectivity of the territories. Thus, a territory is contiguous if there is a subnetwork with one vertex serving as the sink, which receives one unit of flow from every other vertex within the territory.

Constraints $(14)$ ensure that all assigned customers have exactly one predecessor and one successor, and equalities $(15)$ are the flow conservation constraints on the depot for each vehicle $k$ and day $d$.

Person consistency is guaranteed in $(16)$. So, if customer $i$ requires service on days $\alpha$ and $\beta$, then $o_{i,\alpha} + o_{i,\beta} - 2 = 0$, enforcing person consistency by $y_{i,k,\alpha}=y_{i,k,\beta}$. Inequalities $(17)$ and $(18)$ set the service start-times at the customers. So, if customer $j$ is visited right after customer $i$ for some vehicle $k$ and day $d$, then $x_{i,j,k,d} = 1$ and we obtain $s_{i,d} + g_i + t_{i,j} + e_{j,d} = s_{j,d}$ from constraints $(17)$ and $(18)$. Otherwise, the arrival times for the two customers are not related. Inequalities $(17)$ also prevent sub-tours, because they enforce increasing service start-time for customers along a route. 

Constraints $(19)$ enforce that vehicles return to the depot on time. The time windows of the customers are enforced by inequalities $(20)$. Finally, the decision variables' integrality and non-negativity requirements are guaranteed by constraints $(21)$-$(28)$.

\section{Solution Framework}
Based on the algorithm for minimizing the number of vehicles in ConVRPTW proposed in \cite{lespay2021}, we develop an algorithm for reducing the number of territories for TD-DMPVRPTW. The proposed algorithm consists of two main heuristics: the Territory Elimination Heuristic and the Merge Heuristic. 

The Territory Elimination Heuristic starts with an initial solution $R$. In this initial solution, there is one territory for each customer, i.e., each basic unit is a territory on its own. Then, in each Territory Elimination Heuristic's iteration, one territory is eliminated, and the corresponding basic units are merged into other territories by means of the Merge Heuristic. Thus, at the beginning of each iteration of the Territory Elimination Heuristic, we have a complete feasible solution $R$, choose one of the territories and eliminate it, creating a partial feasible solution $\bar{R}$ together with an Ejection Pool ($EP$) consisting of the basic units from the eliminated territory. In the Merge Heuristic, we iteratively modify $\bar{R}$, trying to reaccommodate the basic units from $EP$ in the remaining territories. If the Merge Heuristic is successful, it finishes with $EP$ being empty and $\bar{R}$ being a complete feasible solution. In this case, we update $R$ to be equal $\bar{R}$ and proceed to the following iteration of the Territory Elimination Heuristic. The alternative paths and all the details of these algorithms are described in the following subsections. 

Finally, we reoptimize the final solution obtained with respect to the total travel time of the routing plan and the total compactness of the territories. This reoptimization process consists of sequential applications, the well-known interroute relocation optimizing the territories' compactness, and intraroute 2-opt optimizing the routing plan. These local operators are executed until a local optimum is reached (see \cite{braysy2005a}).

\subsection{Territory Elimination Heuristic for TD-DMPVRPTW}

The Territory Elimination Heuristic initializes the current solution $\hat{R}$ to be equal the incumbent solution $R$, and orders the territories present in $\hat{R}$ according to the ascending total number of {\em active} customer-days over the planning horizon. Then, we iteratively choose each territory in this order and try to eliminate it, reaccommodating its basic units in the remaining territories using the Merge Heuristic. In each iteration of the Territory Elimination Heuristic, we temporarily remove the chosen territory, creating the partial feasible solution $\bar{R}$ and initializing the ejection pool $EP$, and execute the Merge Heuristic on $\bar{R}$ and $EP$. 

If the Merge Heuristic is successful, it terminates with $EP$ being empty and $\bar{R}$ being a complete feasible solution. In this case, we update the current solution $\hat{R}$ to be equal $\bar{R}$. Also, if the number of territories in the current solution $\hat{R}$ is strictly smaller than the number of territories in the incumbent solution $R$, then we update $R$ to be equal $\hat{R}$ and push it onto a stack $S$ used for backtracking. Then, the Territory Elimination Heuristic is reinitiated on the solution $\hat{R}$. However, if the Merge Heuristic fails, it means that, given the restricted number of merges, not all basic units could be accommodated in other territories satisfying the capacity, time window, contiguity, and compactness constraints. In that case, we proceed to the next iteration of the Territory Elimination Heuristic, maintaining the same current solution $\hat{R}$. Finally, when all attempts to eliminate one of the territories in $\hat{R}$ have failed, a backtracking procedure is applied.

The backtracking procedure consists in rolling back to a previous solution with one more territory and continuing to iterate from the point where we left - trying to eliminate the following territories. If an attempt to eliminate all the territories of the previous solution has already been made, another rollback is done. The procedure continues recursively until we arrive at a solution that still has some territories to be checked or the backtracking stack gets empty.

\begin{figure}[th!]
        \centerline{\includegraphics{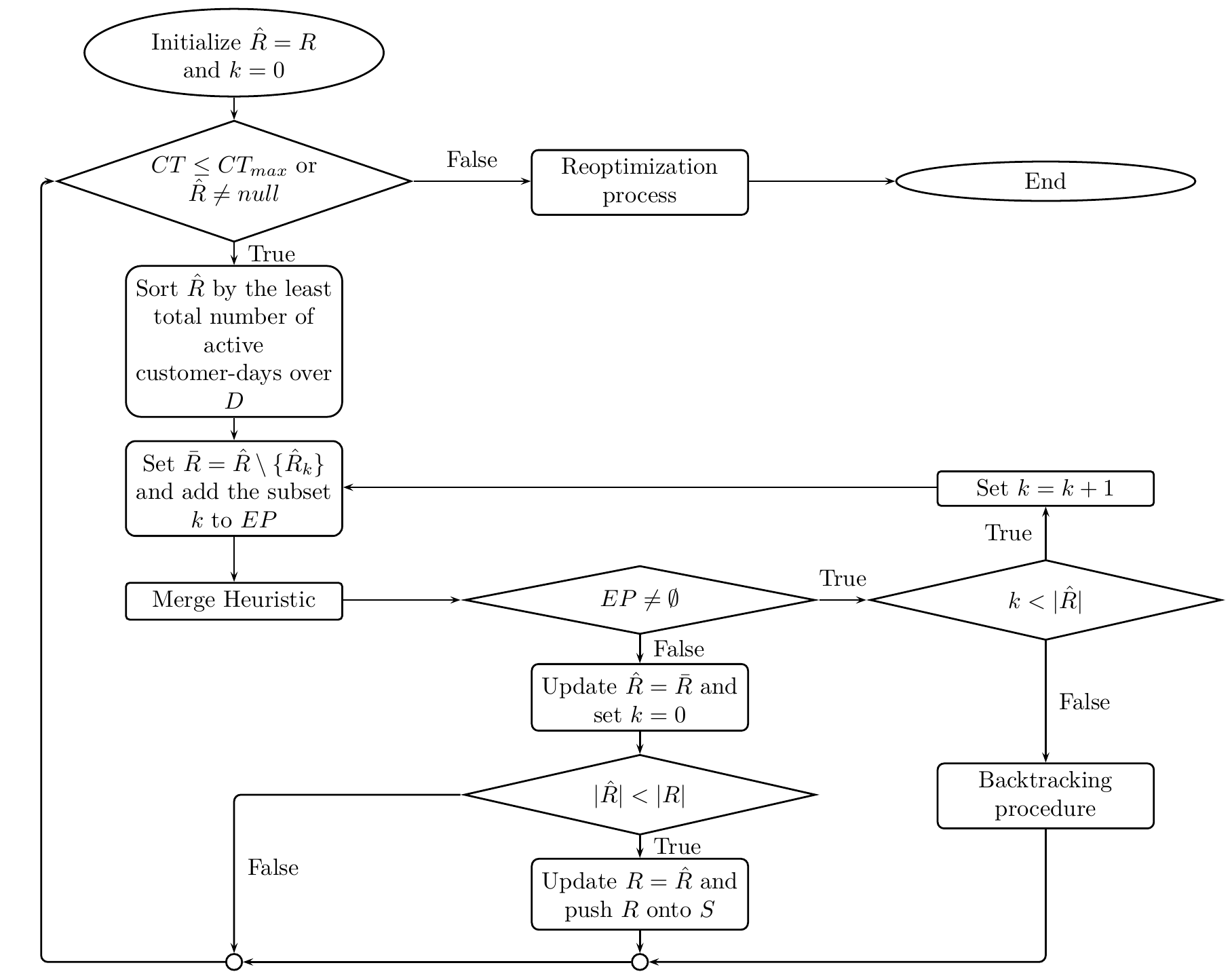}}      
	\caption{Territory Elimination Heuristic for TD-DMPVRPTW\label{fig:E}}
\end{figure}

The backtracking lets to escape from some local optima and explore other search space regions by testing different territory elimination sequences. Due to memory limitations, we implement the backtracking using the stack $S$ with a fixed capacity $\eta$. This limits backtracking from a current solution $\hat{R}$ with $|\hat{R}|$ territories to a previous solution with at most $|\hat{R}| + \eta - 1$ territories. The Territory Elimination Heuristic continues its execution until the total computation time reaches a given limit $CT_{max}$ or the backtracking stack $S$ gets empty, i.e., when all territory elimination sequences available in the backtracking stage have been tested. See Figure \ref{fig:E} for a general overview.

\subsection{Merge Heuristic for TD-DMPVRPTW}
The Merge Heuristic starts with a partial solution $\bar{R}$ and an $EP \neq \emptyset$, provided by the Territory Elimination Heuristic. The heuristic proceeds iteratively, with three stages executed sequentially until merging all basic units from the $EP$ with other territories in $\bar{R}$, or a stopping condition related to the number of merge attempts is reached. In each iteration, a basic unit is selected from $EP$, and an attempt is made to merge it with some territory using the first two stages. If the merge fails in both stages, the basic units' penalty value is incremented by one. Then, in the third stage, the basic unit is merged with a territory selected at random, and some other basic units are ejected, i.e., moved from the respective territory to $EP$, to recover the partial solution's feasibility. 

\begin{figure}[th!]
        \centerline{\includegraphics{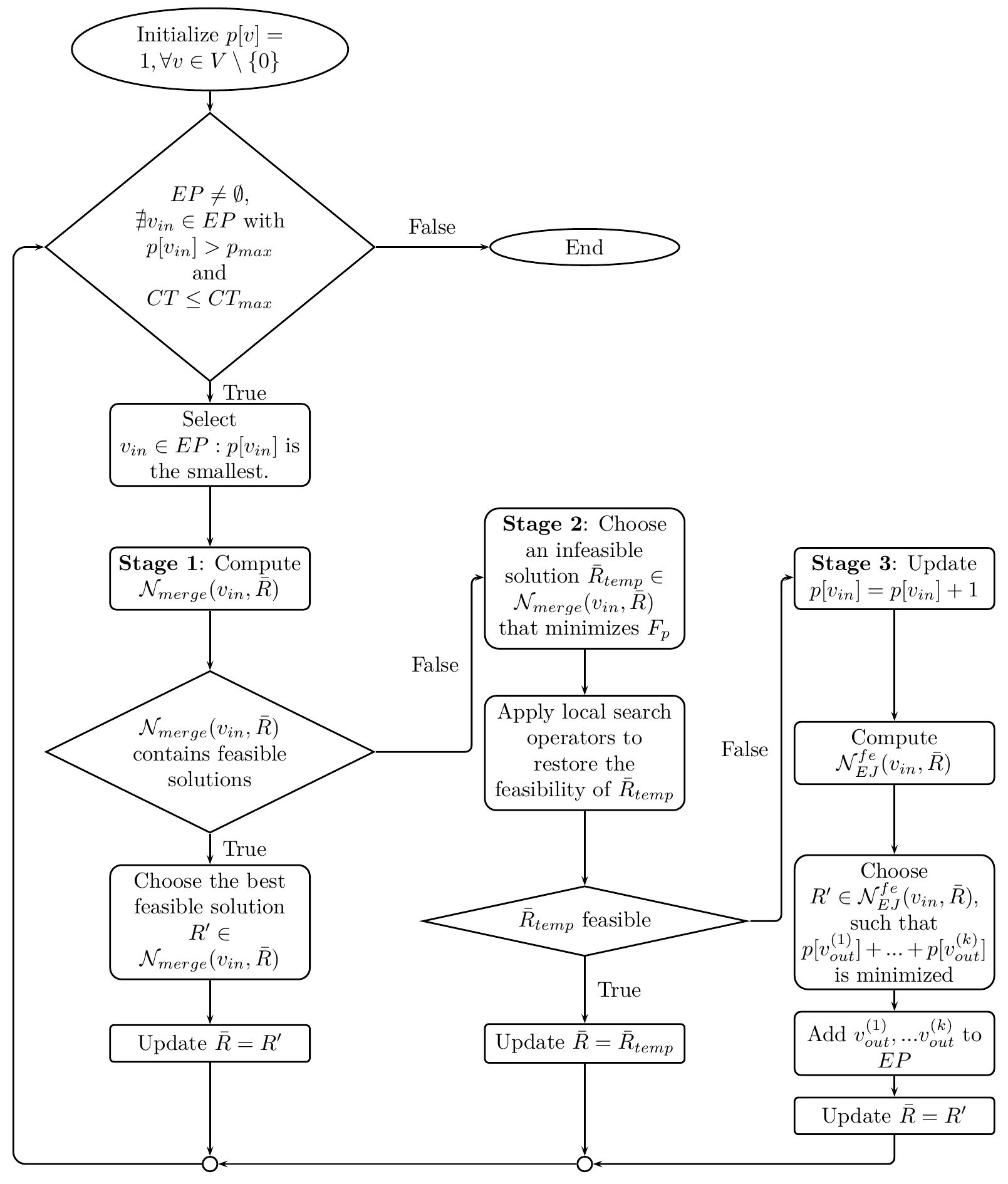}}
	\caption{Insertion Heuristic for TD-DMPVRPTW\label{fig:M}}
\end{figure}

At the beginning, we initialize the penalty value as $p[i] = 1$, $\forall i \in V\setminus \{0\}$. This variable $p$ represents the penalty of the basic unit $i$, which is incremented each time an attempt to merge $i$ with some territory fails at the first two stages of the algorithm. The three stages of the heuristic are applied sequentially until $EP = \emptyset$, at least one basic unit in $EP$ meets the condition $p[i] > p_{max}$, or the total computation time has reached a given limit $CT_{max}$. The second condition considers the number of times a basic unit was merged using stage three and later ejected from its territory. This represents the difficulty in merging a basic unit $i$ with some territory. So when the penalty exceeds $p_{max}$, it is better to let the heuristic explore other search space regions. See Figure \ref{fig:M} for a general overview.

In each iteration of the Merge Heuristic, we select a basic unit $v_{in}$ from $EP$ to merge with some territory. We choose a basic unit $v_{in}$ that satisfies the contiguity constraint and has the lowest penalty value. If there exists at least one basic unit $i$ with penalty value $p[i] > p_{max}$, then we stop the Merge Heuristic. A basic unit $v_{in}$ satisfies de contiguity restriction if there exists at least one territory in the partial solution $\bar{R}$ that has a basic unit adjacent to $v_{in}$. 

In the first stage, we define the set of partial solutions obtained by merging the basic unit $v_{in}$ with each of the territories as $\mathcal{N}_{merge}(v_{in},\bar{R})$. Merging a basic unit with a territory considers all possible insertion positions of the respective customer on the territory's route for each day of the planning horizon. The merge of a basic unit with a territory is feasible only if: (i) the basic unit is adjacent to the territory, ii) the resulting territory satisfies the compactness restriction $\frac{territory\_perimeter}{ \sqrt{territory\_area}} \leq F$, (iii) there exists an insertion position for each day of the planning horizon when the client has positive demand that leads to a feasible route. Recall that $F$ is a parameter of the problem.

 A feasible merge of the basic unit $v_{in}$ with a territory $k$ corresponds to a set of feasible insertions of $v_{in}$ into the route $r_{k,d}$, for each day $d$ when $v_{in}$ has a positive demand. For each day, the best insertion position is computed as in \cite{lespay2021}. If there exist feasible insertions, then the one with the lowest cost is chosen. If $\mathcal{N}_{merge}(v_{in},\bar{R})$ does not contain feasible merges, we move forward to the next stage.

In the second stage, an infeasible merge is chosen from $\mathcal{N}_{merge}(v_{in},\bar{R})$ and temporarily accepted, such that $F_p(R) = \sum _{d\in D} P_{c,d}(\bar{R}) + P_{tw,d}(\bar{R})$ is minimum. $F_p$ is a penalty function defined as the sum of $P_c$ and $P_{tw}$, the penalty terms for the violation of the capacity and time window constraints, respectively. $P_c$ is defined as the total excess of capacities in all routes. $P_{tw}$ is defined as $P_{tw} = \sum_{r=1}^m TW_r$, where $TW_r$ is the sum of time window violations in the route $r$. A series of local search moves, intra-route 2-opt, and inter-route relocation are alternately performed to restore the partial solution's feasibility. If the partial solution's feasibility is restored, we update the partial solution before selecting a new basic unit from $ EP $. Otherwise, we move forward to the next stage.

Finally, in the third stage, since $v_{in}$ could not be merged easily, its penalty value is updated: $p[v_{in}] = p[v_{in}] + 1$. We choose a territory at random, and pick the merge that minimizes the value of $F_p$ - as in the second stage. We define $\mathcal{N}_{EJ}^{fe}(v_{in},\bar{R})$ as the set of feasible partial solutions that are obtained by merging $v_{in}$ with the selected territory, and ``feasibly ejecting'' at most $k_{max}$ basic units, $v_{out}^{(1)},...,v_{out}^{(k)}$ ($k \leq k_{max}$), to restore feasibility. Here also $v_{in}$ itself can be ejected. A basic unit $v_{out}^{(i)}$, $1 \leq i \leq k$, can be ``feasibly ejected'' if if is adjacent to a basic unit in $EP$ or it is adjacent to another basic unit $v_{out}^{(j)}$, $1 \leq j \leq k$, that can be ``feasibly ejected''. This condition prevents the algorithm from ejecting a basic unit that is part of a territory's interior, isolated from other territories by other basic units that are not being ejected - which would make it very difficult to merge it with a different territory afterwards.  

The next solution is selected from $\mathcal{N}_{EJ}^{fe}(v_{in},\bar{R})$ such that the sum of the penalty counters of the ejected basic units, $P_{sum} = p[v_{out}^{(1)}] + ... + p[v_{out}^{(k)}]$, is minimized. Ties are broken by selecting the merge that gives a territory with the minimum ratio $\frac{territory\_perimeter}{\sqrt{territory\_area}}$ (prioritizing the territory's compactness). The ejected basic units are added to $EP$. Then, we update the current partial solution $\bar{R}$ before proceeding to select a new basic unit from $EP$.

\section{Experimental Results}
The proposed algorithm was implemented in Python (see \cite{R07}), using NumPy (see \cite{O15}) and Numba (see \cite{LPS2015}). The proposed MILP model was solved using Gurobi 9.0.1. Both were tested on a set of small instances constructed from Solomon's VRPTW instances, on a PC with Intel Core i7-7700HQ (4 cores @2.8GHz) CPU and 12GB of RAM. The proposed algorithm was also tested on large instances constructed from the company's historical data under study on the same PC. Time and distance matrices were computed using Open Source Routing Machine software (see \cite{LV2011}) based on Open Street Maps data.

\subsection{Instance Characteristics}

\subsubsection{Small TD-DMPVRPTW Instances}
We generate 56 small instances from VRPTW benchmark instances of Solomon\cite{solomon1987}. The original instances are divided into six classes. Classes C1 and C2 have clustered geographical distribution of customers, in classes R1 and R2 the geographical distribution is randomly uniform, and classes RC1 and RC2 have a mix of both geographically uniform and clustered customers. Classes R1, C1 and RC1 have narrow time windows and small vehicle capacities of 200, so the number of customers served by a single vehicle is small. Classes R2, C2 and RC2 have wide time windows over longer scheduling horizons and large vehicle capacities of 700 for C2 and 1000 for R2 and RC2, so the solutions can have comparatively fewer vehicles. All of the classes consider 100 customers and a single depot.

From Solomon's instances, we construct small TD-DMPVRPTW instances, as in \cite{lespay2021}. We consider only the first ten customers and keep without modifying the customers' demands, coordinates, time windows, and service times. We consider a planning horizon of five days. The service frequency, i.e., the probability that a customer requires service on a given day, was set to 70\%. We reduced the capacity of vehicles by half.

\subsubsection{Large TD-DMPVRPTW Instances}
A set of fifteen large instances was constructed from the company's historical monthly demand data from January 2018 until March 2019. Each one of them is characterized by a different number of customers and days. The vehicle fleet is homogeneous, with vehicles of 5 tons each.

In Table, \ref{tab:inst} we present a brief description of these instances. They are characterized by a long planning horizon of 23 days, with 1784 customers and a total of 8129 orders, on average. Moving from one month to the following one, on average, there are 11.5\% of new customers. We highlight that, on average, only 19.2\% of the customers are active each day - a low proportion of the total number of customers. This situation makes it particularly difficult to design territories compatible with daily routing plans.

\begin{table}[th!]
{\scriptsize
\caption{Monthly instances}\label{tab:inst}
\begin{tabular*}{\hsize}{@{}@{\extracolsep{\fill}}llccccccccc@{}}\toprule
&&&&\multicolumn{2}{c}{\textbf{Customers}}&\multicolumn{4}{c}{\textbf{Active customers per day [\%]}}\\
\cmidrule(r){5-6}\cmidrule(r){7-10}
Id&Instance	&Days &Orders  &Current &New[\%] &Mean &Std. &Min. &Max. \\
\toprule

1	&January2018	&26		&10142	&2079	&-		&18.8	&4.1	&7.6	&24.6 \\
2	&February2018	&21		&7857	&1958	&10.0	&19.1	&4.0	&12.6	&25.8 \\
3	&March2018		&18		&6526	&1840	&12.1	&19.7	&4.2	&9.6	&25.7 \\
4	&April2018		&25		&8681	&1892	&16.1	&18.4	&3.7	&11.5	&23.6 \\
5	&May2018		&19		&6832	&1776	&9.7	&20.2	&4.5	&8.0	&26.1 \\
6	&June2018		&26		&9175	&1899	&15.3	&18.6	&3.5	&12.8	&23.1 \\
7	&July2018		&26		&8652	&1772	&8.2	&18.8	&4.6	&8.2	&26.1 \\
8	&August2018		&27		&9476	&1859	&12.6	&18.9	&3.4	&12.5	&25.4 \\
9	&September2018	&22		&7284	&1719	&7.6	&19.3	&5.0	&7.9	&26.6 \\
10	&October2018	&24		&7451	&1698	&11.3	&18.3	&4.2	&6.9	&24.3 \\
11	&November2018	&24		&7277	&1641	&12.0	&18.5	&4.5	&8.2	&26.1 \\
12	&December2018	&25		&8180	&1678	&13.1	&19.5	&4.4	&10.6	&27.8 \\
13	&January2019	&26		&8603	&1714	&12.8	&19.3	&2.8	&13.1	&23.5 \\
14	&February2019	&24		&8051	&1622	&8.0	&20.7	&3.0	&15.4	&25.8 \\
15	&March2019		&24		&7751	&1625	&12.0	&19.9	&3.1	&14.5	&25.5 \\
	&Average		&23.8	&8129.2	&1784.8 &11.5	&19.2	&3.9	&10.6	&25.3 \\
\bottomrule
\end{tabular*}
}
\end{table}

\subsection{Experimental Settings}
The parameters to be adjusted in our algorithm correspond to the Territory Elimination Heuristic and the Merge Heuristics.

Most parameters of the algorithm were set the same as the values used in \cite{lespay2021}, with the maximum computational time $CT_{max} = \{1,10,60\}$, in minutes, and $k_{max} = 3$.  Additionally, we chose values for $\eta$, $p_{max}$, $F$. Different alternatives for these parameters were tested, but we report those that offer the best results in terms of the average number of vehicles, and assure a good level of compactness of territories. In particular, we set $\eta=5$, $p_{max}=5$, and $F=10$. Other parameter settings of the algorithm are available in \cite{lespay2021}.

The values of these parameters were maintained unchanged for all the instances. Consistently good performance of these parameter values makes our algorithm easy to implement since no per-instance parameter tuning is necessary.

\subsection{Results for Small ConVRPTW Instances}

We evaluate the TD-DMPVRPTW algorithm's performance by comparing the solutions obtained by TD-DMPVRPTW with the optimal solutions found by Gurobi, with a time limit of 1 hour. We set the TD-DMPVRPTW to stop when the optimal solution reported by Guroby is found.

The results of the comparison are reported in Table \ref{tab:R_small}. All the instances were solved optimally in 1 hour. We present the average result for every class. In the first four columns, we report the optimal solution with respect to the average number of vehicles (ANV), the average compactness ratio of the territories designed (ACR), the average total travel time (ATT), and the average total computational time (ACPU) in seconds. In the following four columns, we report the TD-DMPVRPTW algorithm's solution with respect to ANV, ACR, ATT, and ACPU. 

We can see that the number of vehicles obtained by the TD-DMPVRPTW algorithm is the same as the one obtained with the MILP model using Gurobi for all instance classes, but with a much lower computing time.

\begin{table}[th!]
{\scriptsize
\caption{Results for small TD-DMPVRPTW instances}\label{tab:R_small}
\begin{tabular*}{\hsize}{@{}@{\extracolsep{\fill}}lcccclcccc@{}}\toprule
&\multicolumn{4}{c}{\textbf{Gurobi}}&\multicolumn{4}{c}{\textbf{TD-DMPVRPTW}}\\

\cmidrule(r){2-5}\cmidrule(r){6-9}
Instance	&ANV &ACR &ATT &ACPU	 &ANV &ACR &ATT &ACPU\\
  			&  	 &	 &    &[s]	 &    &	  &    &[s]
 \\\midrule
C1		&1.9	&4.8    &9310.5		&6.6	&1.9	 &4.4 	&6847.9		&0.03\\
C2		&1.0	&5.2    &14129.8	&3.1	&1.0	 &5.2 	&13952.2	&0.01\\
R1		&2.4	&5.4    &1825.5		&324.7	&2.4	 &5.0  	&1860.8		&0.01\\
R2		&1.0	&5.9    &3369.0		&241.3	&1.0	 &5.9 	&3242.1		&0.01\\
RC1		&2.9	&5.5    &2236.6		&289.1	&2.9	 &4.8 	&2214.9		&0.01\\
RC2		&1.0	&4.9    &3243.2		&159.5	&1.0	 &4.9 	&3159.3		&0.01\\
Avg.	&1.7	&5.3    &5685.7		&170.7	&1.7	 &5.0 	&5212.9		&0.01\\
\bottomrule
\end{tabular*}
}
\end{table}

\subsection{Results for Large TD-DMPVRPTW Instances}

The territories generated by our heuristic for TD-DMPVRPTW are compared with the company's current customer-to-driver assignments. With the assignments, it is often impossible to find a daily routing plan that respects the customers' time windows. Moreover, they do not correspond to contiguous territories. We evaluate the performance of the current territory design of the company in terms of the number of vehicles used for executing the deliveries (NV), the total travel time resulting from the daily routing plans (TT) in hours, the percentage of visits for which the time windows are not respected (PTW), and the total lateness of deliveries with respect to time windows as a proportion of TT (LTW).

\begin{landscape}
\begin{table}[th!]
{\scriptsize
\caption{Results for monthly instances}\label{tab:R_large}
\begin{tabular*}{\hsize}{@{}@{\extracolsep{\fill}}lccccccccccccccccc@{}}\toprule
&\multicolumn{4}{c}{\textbf{Current Solution}}&\multicolumn{8}{c}{\textbf{TD-DMPVRPTW}} \\

\cmidrule(r){2-5} \cmidrule(r){6-17}
&&&&&  1min. &&&& 10min. &&&& 60min. &\\

\cmidrule(r){2-5}\cmidrule(r){6-9}\cmidrule(r){10-13}\cmidrule(r){14-17}
Id. &NV &TT &PTW &LTW &NV &$\Delta_{TT}$ &ACR&CPU &NV &$\Delta_{TT}$ &ACR&CPU &NV &$\Delta_{TT}$ &ACR&CPU &$\Delta_{NV}$\\
    &   &[h]&[\%]&[\%]&   &[\%]          &   &[min]&   &[\%]          &   &[min] &   &[\%]          &   &[min]     &[\%]\\
\toprule
1	&22	  &5024.5 &16.4	&12.5  &20	 &15.3	&4.9 &2.2  &18	 &17.6	&5.3 &11.4 	&17	  &19.0	&5.4 &31.3  &22.7\\
2	&22	  &3901.1 &8.3	&6.0   &17	 &15.7	&5.4 &3.0  &17	 &16.0	&5.4 &11.4 	&16	  &18.1	&5.6 &45.2  &27.3\\
3	&23	  &3278.8 &7.1	&6.5   &21	 &11.4	&5.0 &2.7  &20	 &13.5	&5.1 &11.1 	&17	  &16.0	&5.6 &61.5  &26.1\\
4	&22	  &4416.1 &6.0	&4.2   &20	 &10.3	&5.2 &2.7  &19	 &12.1	&5.2 &11.4 	&18	  &14.2	&5.2 &61.4  &18.2\\
5	&22	  &3468.5 &9.8	&7.7   &19	 &14.5	&4.9 &1.9  &19	 &14.5	&4.9 &11.3 	&18	  &15.2	&5.5 &61.0  &18.2\\
6	&22	  &4713.9 &9.3	&6.3   &19	 &13.8	&5.0 &2.1  &19	 &13.8	&5.0 &11.1 	&18	  &15.3	&5.2 &61.0  &18.2\\
7	&22	  &4691.8 &10.3	&6.4   &18	 &16.2	&5.3 &2.2  &17	 &18.0	&5.4 &11.1 	&17	  &17.5	&5.4 &21.4  &22.7\\
8	&22	  &4858.7 &3.9	&2.1   &18	 &15.8	&5.1 &2.7  &17	 &17.2	&5.1 &3.3 	&16	  &18.6	&5.6 &60.4  &27.3\\
9	&23	  &3965.9 &6.1	&4.4   &17	 &16.2	&5.7 &2.1  &17	 &16.2	&5.7 &10.9 	&16	  &17.6	&5.8 &31.1  &30.4\\
10	&22	  &3979.6 &6.1	&3.7   &16	 &17.5	&5.2 &1.9  &16	 &16.5	&5.5 &10.9 	&16	  &17.5	&5.2 &60.9  &27.3\\
11	&21	  &4036.7 &4.7	&2.6   &17	 &17.6	&5.0 &1.4  &16	 &19.0	&5.3 &10.5 	&16	  &19.0	&5.3 &14.5  &23.8\\
12	&23	  &4330.3 &5.7	&2.9   &17	 &17.8	&5.3 &2.4  &17	 &16.8	&5.3 &11.2 	&16	  &18.4	&5.7 &52.1  &30.4\\
13	&21	  &4527.7 &7.2	&3.7   &17	 &16.8	&5.3 &2.1  &17	 &16.8	&5.3 &11.2 	&16	  &18.4	&5.4 &61.4  &23.8\\
14	&21	  &4267.2 &9.3	&5.7   &16	 &19.3	&5.0 &2.1  &16	 &18.8	&5.0 &10.9 	&16	  &19.0	&5.1 &61.1  &23.8\\
15	&21	  &4036.6 &4.4	&2.5   &15	 &19.8	&5.1 &1.9  &15	 &19.8	&5.1 &10.7 	&15	  &19.8	&5.1 &49.9  &28.6\\
Avg.&21.9 &4233.2 &7.6	&5.1   &17.8 &15.9	&5.2 &2.2  &17.3 &16.4	&5.2 &10.6 	&16.5 &17.6	&5.4 &48.9  &24.6\\
\bottomrule
\end{tabular*}
}
\end{table}
\end{landscape}

In table \ref{tab:R_large}, we describe the performance of the company's current operation on the monthly instances, in contrast to the results generated by our algorithm for TD-DMPVRPTW. In the first four columns of this table, we report NV, TT, PTW, and LTW of the company's current customer-to-driver assignments. In the next columns, we report the results for the territories obtained by applying the proposed TD-DMPVRPTW algorithm, which are evaluated by NV, the percentage of improvement of the total travel time $\Delta_{TT}$, the average compactness ratio of territories (ACR), and the total computation time (CPU) in seconds. The last column gives the percentage of improvement in the number of vehicles $\Delta_{NV}$ between the company's current solution and the best results obtained with the TD-DMPVRPTW algorithm. We present the results obtained using maximum computation times of 1, 10, and 60 minutes for our algorithm. 
The complete program's execution time is equal to the chosen configuration time plus the time spent for setting the initial solution and the final re-optimization heuristic.

In Table \ref{tab:R_large}, in the current territory design, the number of vehicles required for executing the deliveries ranges from 21 to 23 vehicles, there exists, on average, 7.6\% of customers for which the time windows are not respected, and the total lateness of deliveries with respect to time windows as the proportion of TT is 5.1\%, on average. The total travel time required for the complete routing plan is, on average, 4233.2 hours. Our algorithm reaches good solutions already with the setting of 1 minute. Using the setting of 10 minutes leads to a further reduction of one vehicle on instances 3,7,8,11, and two vehicles for instance 1. A setting of 60 minutes reduces one vehicle on almost all of the instances and three vehicles on instance 3. The total travel time decreases, on average, by at most 1.2 percent points from one time setting to another. The maximum percentage of improvement in the number of vehicles is, on average, 24.6\%. In brief, the company's routing plan is improved, reducing the number of vehicles required for the deliveries, on average, in five vehicles, the total travel times are improved by 17.6\%, on average, and the customers are serviced within their time windows. Finally, we can note that the territories reach a good average compactness ratio, with values of at most 5.6.   

Notice that, since the total travel time (TT) is not part of the objective function that we are optimizing, in some cases, more computational time leads to higher values of this attribute. This is partly due to the restrictions imposed by the time windows, where reducing the number of vehicles may force longer routes - like in the instances number 10, 12, and 14, when augmenting the computation time from 1 to 10 minutes, and in the instance number 7, when increasing the computation time from 10 to 60 minutes. But it may also be due to certain random characteristics of our algorithm. Similar cases can be observed for the average compactness ratio (ACR).

\subsection{Analysis of the TD-DMPVRPTW algorithm for Operational Management}

In this subsection, we provide some managerial insights by evaluating the territories computed by the proposed heuristic in the context of the company's operational reality under study. With a fixed customer-to-driver assignment, the company tries to satisfy the customers' demand for every unknown operational day. However, because of the variability of customers' demand and activity, the company is forced to implement ad-hoc changes to the assignment on individual days. For instance, when a driver has active customers with a demand that exceeds the vehicle's capacity on a particular day, some of these customers are reassigned to other drivers. In some cases, there may even be orders not fulfilled at all. Therefore, this situation generates delays in the deliveries and reduces the consistency of the service.

This subsection's general idea is to use the company's historical data from one month to compute customer-to-driver assignments through TD-DMPVRPTW and evaluate the performance of the obtained territories in daily operational routing during the following month. The analysis begins with a TD-DMPVRPTW solution obtained from solving each monthly instance, respectively. Then, we analyze their daily performance during the following month. Our objective is to evaluate the resulting daily routing plans' feasibility.

The general rule we evaluate is to respect the customer-to-driver assignment for old customers and to assign each new customer to the driver that serves its nearest old customer. Such a rule can be easily implemented by a human truck dispatcher, without any need for computational support.

In table \ref{tab:R_operacional}, we present the performance of the territories obtained by TD-DMPVRPTW when they are used for daily operational routing during the following month. We report the total number of infeasible new customers (TIC) - for these customers; it was not possible to serve them with the vehicle assigned to the nearest old customer; the total number of days with infeasible customers (TID), the total demand of infeasible customers (IAC) in [kg], and the average lateness of deliveries with respect to time windows of infeasible customers (IATW) in [h]. In the table, we can observe that the operational solutions constructed from the TD-DMPVRPTW are feasible most of the time, resulting infeasible only on instances 9,12, and 13, with a low proportion of infeasible customers and days. Therefore, our solutions of TD-DMPVRPTW can be used to easily create operational routing plans that will be feasible in a significant proportion of the time.

\begin{table}[h]
{\scriptsize
\caption{Results for operational following month experiments from monthly instances}\label{tab:R_operacional}
\begin{tabular*}{\hsize}{@{}@{\extracolsep{\fill}}llcccc@{}}\toprule

Id	&Following month &TIC &TID  &IAC  &IATW \\	
	&			     &    &     &[kg] &[h]	    
 \\\midrule
1	&February2018	&0	 &0	    &0.0	&0.0\\
2	&March2018		&0	 &0	    &0.0	&0.0\\
3	&April2018		&0	 &0	    &0.0	&0.0\\
4	&May2018		&0	 &0	    &0.0	&0.0\\
5	&June2018		&0	 &0	    &0.0	&0.0\\
6	&July2018		&0	 &0	    &0.0	&0.0\\
7	&August2018		&0	 &0	    &0.0	&0.0\\
8	&September2018	&0	 &0	    &0.0	&0.0\\
9	&October2018	&1	 &1	    &0.0	&0.1\\
10	&November2018	&1	 &1	    &1268.1	&0.0\\
11	&December2018	&0	 &0	    &0.0	&0.0\\
12	&January2019	&3	 &3	    &207.4	&0.0\\
13	&February2019	&1	 &1	    &1206.2	&0.0\\
14	&March2019		&0	 &0	    &0.0	&0.0\\
    &Avg.			&0.4 &0.4	&191.6	&0.004\\
\bottomrule
\end{tabular*}
}
\end{table}

\section{Conclusions and Future Work}

This study introduces the Territory Design for Dynamic Multi-Period Vehicle Routing Problem with Time Windows (TD-DMPVRPTW). This problem arises in the context of a real-world application at a food company's distribution center, in which the main objective is to design the minimum number of contiguous and compact territories for delivery of orders from a depot to a set of customers, with time windows, over a multi-period planning horizon. 

We develop a mixed-integer linear programming (MILP) model and a heuristic algorithm for the TD-DMPVRPTW. We compare both approaches on small instances from the literature. We show that the heuristic can reach the optimal solutions, reported by Gurobi Optimizer based on the MILP formulation, in a small fraction of the solver's computational time. For large instances, based on the company's real-life data under study, the proposed algorithm reports territories that generate routing plans with a number of vehicles significantly lower than those currently used by the company within reasonable computational times. 

Additionally, we test the territories generated by our algorithm in a past month as the basis for creating operational daily routing plans for the following month, with the customer activity and demand known only one day in advance. Using a simple rule of respecting the customer-to-driver assignment for the old customers and assigning each new customer to the driver that serves its nearest old customer, we obtain feasible solutions most of the time. This confirms the high flexibility of the territories generated by the algorithm.

Due to the TD-DMPVRPTW algorithm's effectiveness in minimizing the number of territories (vehicles), it can be a valuable tool to create customer-to-driver assignments that correspond to compact geographical zones that do not overlap. This is a useful characteristic for the company's management, for the ease of establishing service contracts with third party transportation companies based on general territories and not on particular sets of clients. This setting makes it easier to deal with high variability in client activity. Moreover, it is possible to define geofences delimiting the trucks' normal operation zones to raise alerts when a deviation is detected (based on real-time GPS tracking) - thus improving the drivers' security trucks and the merchandise.

In future research, we plan to improve the algorithm's efficiency to find good solutions to larger instances, like the historical data of a whole year of past operations. Due to contiguity constraints, it is difficult to escape from local optima and reduce the number of territories exploring neighborhood structures. For this reason, we will explore some possible strategies to embed our algorithm in a multistart framework to improve the efficacy of exploration of the search space due to stronger diversification strategies.

\section*{Acknowledgements}
H.L. gratefully acknowledges financial support from ANID + PAI/Concurso Nacional Tesis de Doctorado en el Sector Productivo, 2017 + Folio T7817120007.
K.S. gratefully acknowledges financial support from Programa Regional STICAMSUD 19-STIC-05.

\bibliographystyle{acm}
\bibliography{references2}

\end{document}